\documentclass[letterpaper, 10 pt, conference]{ieeeconf}  % Comment this line out if you need a4paper

\IEEEoverridecommandlockouts                              % This command is only needed if 
\usepackage{graphicx} % for pdf, bitmapped graphics files
\usepackage{authblk}
\usepackage{placeins}
\usepackage{amsmath}
\usepackage{float}

\title{\LARGE \bf
BRAVE-6D: Benchmark for Robotic Active Vision in 6DOF Pose Estimation
}

\author{Philipp Ausserlechner$^{1,*}$, Bernhard Neuberger$^{1,*}$, \\Alessandro Scherl$^{2}$, Michael Schebek$^{2}$, Stefan Thalhammer$^{2}$, Markus Vincze$^{1}$ \\
\thanks{$^{1}$ Automation and Control Institute, TU Wien, 1040 Vienna, Austria. \{ausserlechner, neuberger, vincze\}@acin.tuwien.ac.at}
\thanks{$^{2}$ Department of Industrial Engineering, UAS Technikum 1200 Vienna, Austria
\{alessandro.scherl, michael.schebek, stefan.thalhammer\}@technikum-wien.at}
\thanks{$*$ These authors contributed equally to this work.}
\thanks{This work was supported by the EU-program EC Horizon 2020 for
Research and Innovation under grant agreement No. 101017089, project
TraceBot and grant agreement No. 101120823, project MANiBOT.}
}

\begin{document}

\maketitle
\thispagestyle{empty}
\pagestyle{empty}

%%%%%%%%%%%%%%%%%%%%%%%%%%%%%%%%%%%%%%%%%%%%%%%%%%%%%%%%%%%%%%%%%%%%%%%%%%%%%%%%
\begin{abstract}
% Detecting and grasping small objects is an open challenge in robotics. The natural approach is to use active vision to come closer, but the comparison of works is difficult due to the interaction with the scene. We propose a benchmark to evaluate robot active vision systems for the task of object pose estimation as first step to grasp objects, BRAVE-6D. The idea is to exploit view synthesis methods based on Gaussian Splats (3DGS) with 6D pose annotations in a setting to freely move the robot. We provide a baseline solution to find small objects capable of navigating scenes and estimating object poses.

Detecting and grasping small objects remains a significant challenge in robotics. 
Active vision, where the robot moves closer to the object, is an intuitive solution, yet comparing approaches on common ground is difficult since identical physical scene setups are required.
Hence, we introduce BRAVE-6D, a benchmark designed to evaluate robotic active vision systems for object pose estimation, a crucial first step in grasping objects. 
BRAVE-6D leverages view synthesis based on Gaussian Splats (3DGS) to provide scenes and tools for benchmarking active vision systems.
We show baseline solutions performing visual servoing within the scene and accurately estimating the poses of small objects.

%We also provide baseline solutions for navigating scenes and accurately estimating the poses of small objects.
%Mobile robots rely on 6D object pose estimation for manipulation tasks, but standard benchmarks evaluate static images, failing to reflect real-world challenges where robots can adjust their views.
%Consequently, there is a need to evaluate agent systems that adapt their perspectives for accurate pose estimation.
%We present \textbf{BRAVE-6D}, a dataset and benchmark for robotic vision tasks that leverages state-of-the-art novel view synthesis methods to provide an active vision challenge.
%Scenes are presented as 3D Gaussian Splats (3DGS) with 6D object pose annotations, and an interface allows agent/camera control to reach desired viewpoints.
%Small objects were chosen as a challenging, industry-relevant use case since they occupy few pixels in the sensor.
%An active vision approach allows moving closer to the object, obtaining more sensor information for precise 6D pose estimation.
%We provide two baseline solutions, a supervised trained and a zero-shot agent, capable of navigating scenes and estimating object poses.
%In summary, \textbf{BRAVE-6D} introduces a new paradigm for evaluating and benchmarking active vision systems for 6D pose estimation, addressing the challenges of small objects.
\end{abstract}

%%%%%%%%%%%%%%%%%%%%%%%%%%%%%%%%%%%%%%%%%%%%%%%%%%%%%%%%%%%%%%%%%%%%%%%%%%%%%%%%
\section{INTRODUCTION}
%Was ist das Problem/Motivation?
A significant challenge in robotic object manipulation is handling small objects, defined by \cite{peng2022small} as less than 32x32 pixels.
Positioning the sensor closer to the object is the logical solution, feasible in scenarios where agents actively select their viewpoint. 
However, state-of-the-art 6D pose estimation benchmarks provide static images~\cite{hodavn2020bop}, therefore not reflecting realistic scenarios.

Active perception approaches~\cite{bajcsy1988active, aloimonos1988active, ilyas2021robot} that adjust the camera pose based on the observation show improved detection and pose estimation. This results in higher success rates for object grasping~\cite{liu2022collaborative, natarajan2021aiding}.
However, due to the dependency of the scene setup, benchmarking such systems is cumbersome and often infeasible.

%Previous active vision datasets like Ammirato et al.~\cite{ammirato2017dataset} capture image data using a mobile robot navigating through indoor scenes, positioning the camera at points on a grid, thus offering a discrete action space for the agent system.
%On the other hand, the Robotic Vision Scene Understanding Challenge (RVSU) builds on a customizable NVIDIA Omniverse and Isaac Sim simulation~\cite{hall2022benchbot}, resulting in a continuous action space, but limiting realism and requiring high computational expenses.
Previous active vision datasets like Ammirato et al.~\cite{ammirato2017dataset, ammirato2018active} capture image data with a mobile robot navigating indoor scenes and positioning the camera on a grid, offering a discrete action space. 
The Robotic Vision Scene Understanding Challenge (RVSU) uses NVIDIA Omniverse and Isaac Sim for simulation~\cite{hall2022benchbot}, providing a continuous action space but limiting realism and requiring high computational resources.

\begin{figure}[t!]
  \centering
  \includegraphics[width=0.46\textwidth]{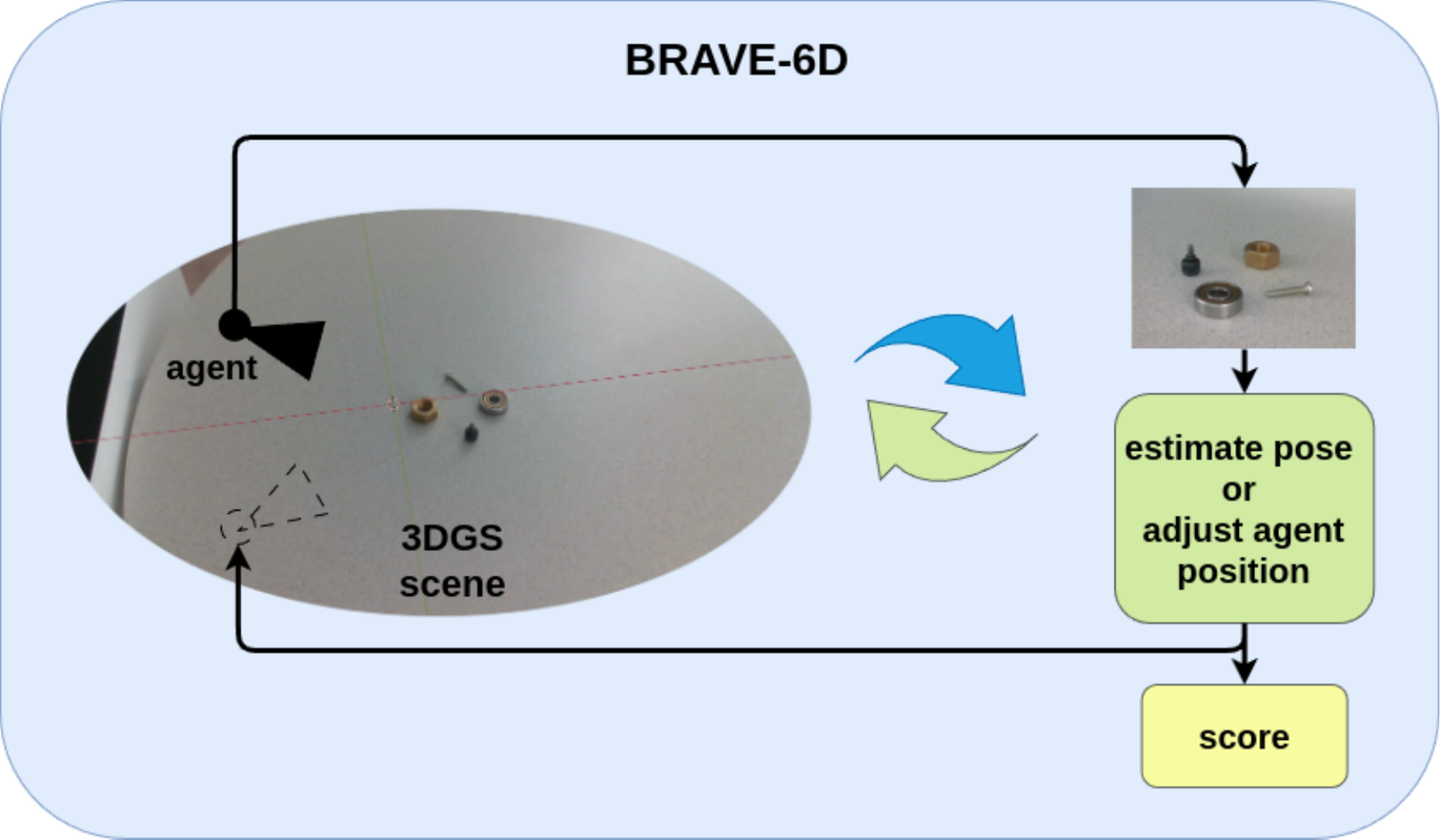}
  \caption{\textbf{BRAVE-6D}Active vision benchmark for 6D pose estimation.}
  \label{fig:teaser}
\end{figure}

To overcome these limitations, we propose \textbf{BRAVE-6D} - a novel type of benchmark where the agent is free to move continuously and actively select viewpoints to achieve the task, e.g., finding an object and going closer to a small item. 
The idea of \textbf{BRAVE-6D} is to offer a continuous, realistic, and lightweight scene representation, enabled by 3D Gaussian Splatting (3DGS)~\cite{kerbl20233d}, with annotated 6D object poses from four categories: industry, toys, medical lab equipment, and office supplies.
We collected the data with a robotic setup to provide uniform view coverage through all scenes.
To evaluate the quality of 3DGS, 10\% of the views are used as a holdout set.
A ROS interface is provided, enabling agents to navigate the 3DGS scene and change its viewpoint.
Additionally, we present two baseline agents for \textbf{BRAVE-6D}.
A zero-shot one, based on CNOS~\cite{nguyen2023cnos} detections and ZS6D~\cite{ausserlechner2023zs6d} pose estimations.
A supervised one, based on YOLO~\cite{redmon2016you} detection and the state-of-the-art pose estimation method GDR-Net~\cite{wang2021gdr}.
The dataset includes a comprehensive training set with synthetic rendered images and ground truth annotations.
Our contributions to active vision, robotics, and 6D object pose estimation are as follows:
\begin{itemize}
    \item \textbf{BRAVE-6D}, an interactive object pose estimation dataset consisting of 3DGS scenes and annotated small objects from four industry-relevant categories.
    \item  A ROS interface for real-time visual servoing and corresponding view generation.
    \item An empirical evaluation of the influence of sensor distance on object pose estimation accuracy.
    \item Two baseline solutions, a supervised trained and a zero-shot agent, capable of visual servoing within the scenes and estimating object poses.
\end{itemize}
In summary, \textbf{BRAVE-6D} enables commanding an interactive agent/camera and offers a realistic and reproducible benchmark for difficult robotic tasks such as handling small objects.
%============================================================
% more here
\begin{figure*}[ht!]
  \centering
  \includegraphics[width=\textwidth]{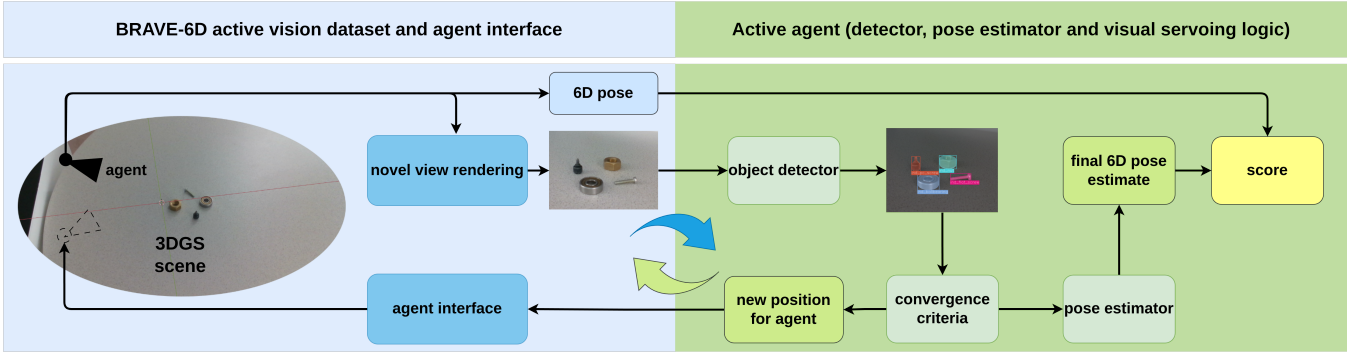}
  %\vspace{-15pt}
  \caption{\textbf{Overview of BRAVE-6D.} The diagram illustrates the two main contributions of our paper: (1) the active vision dataset and (2) the active agent for evaluation. The left side visualizes the active vision dataset, which includes a 3DGS (3D Gaussian Splatting) scene with an interface for positioning the agent and a view renderer that returns the view and the object pose in camera coordinates based on the agent's position. The right side depicts the proposed active agent logic for determining an optimal agent position and estimating the 6D object pose in camera coordinates.}
  \label{fig: pipeline}
\end{figure*}

\FloatBarrier

\section{BRAVE-6D}
Figure~\ref{fig: pipeline} gives an overview of \textbf{BRAVE-6D} and illustrates the feedback loop between the agent and the interactive dataset.
We utilize 3D Gaussian Splatting (3DGS) for scene representation due to its capability for real-time view synthesis.
The agent interacts with the scene through the ROS interface, deciding whether to move to a new position or to estimate the 6D object pose in camera coordinates.

\subsection{Data Collection}
The 3DGS scenes were collected using a camera mounted on a robotic arm, ensuring precise knowledge of the camera poses. 
A 3DGS scene consists of ellipsoids, characterized by a 3D position, opacity, anisotropic covariance, and spherical harmonic coefficients to capture view-dependent color variations. 
These parameters are optimized during the training of the 3DGS model, with the number of ellipsoids determined by an adaptive control algorithm.
The loss function used for training, shown in Equation~\ref{eq: gaussian_loss}, combines a pixel-wise $\mathcal{L}_1$ norm and the structural similarity index loss term $\mathcal{L}_{D-SSIM}$ to measure holistic image similarity.
% \vspace{-2mm}
\begin{align} 
\mathcal{L} = (1- \lambda)\mathcal{L}_1 + \lambda \mathcal{L}_{D-SSIM} 
\label{eq: gaussian_loss} 
\end{align}

% \vspace{-2mm}
We capture images at four levels: 20~\text{cm}, 30~\text{cm}, 40~\text{cm} and 50~\text{cm}, with approximately 90 images per distance. 
This approach facilitates accurate view synthesis from both close and far perspectives.
Object poses are annotated on a scene level with 3D-DAT~\cite{suchi20233d}. 
The dataset includes four objects per category (industry, household, laboratory, and toys).

\subsection{Agent Interface and Evaluation}
The ROS interface allows the agent to move within the 3DGS scene, simulating the behavior of a robot interacting with its environment. 
Retrieving and setting the camera pose and velocity, is possible, constrained only by annotated collision boxes.
The agent detects objects and decides whether to change position or estimate the pose. 
If the view is insufficient, it sends the desired camera position to the ROS interface.
\textbf{BRAVE-6D} returns the corresponding synthesized view and ground truth 6D object pose in the current frame.
Once a satisfactory position is reached, as determined by the convergence criterion, the agent estimates the 6D object pose.
%We evaluate two metrics: the Average Recall (AR) score from the BOP challenge~\cite{hodavn2020bop} for pose estimation accuracy, and the number of agent position adjustments.
%The latter metric directly correlates with the visual servoing effort in a real-world scenario. 

\subsection{Evaluation Metrics}
The evaluation will consist of two metrics: the typical Average Recall (AR) score from the BOP challenge~\cite{hodavn2020bop} for pose estimation accuracy, and the number of agent position adjustments.
The latter metric directly correlates with the visual servoing effort in a real-world scenario. 
%This approach provides not only a new benchmarking procedure for mobile computer vision agents but also a benchmark for visual servoing tasks.

\subsection{Zero-Shot Active Vision Pipeline}
To establish a baseline for BRAVE-6D, we provide a zero-shot active vision pipeline that includes CNOS as a detector and ZS6D as a subsequent pose estimation module. 
This demonstrates that no training is necessary to estimate the 6D pose of challenging objects if an optimal viewpoint is selected.

\section{Conclusion}
\textbf{BRAVE-6D} is not merely a dataset; it represents a new strategy for dataset creation and benchmarking for active perception, enables comparing methods, and aims to foster the development of new approaches that outperform the presented baseline solutions.

Future work can build on our strategy to create benchmarks for scenarios like object searching and extend the 3DGS toward physics properties to enable a comparison of mobile robotic algorithms.

\addtolength{\textheight}{-12cm}   % This command serves to balance the column lengths

%%%%%%%%%%%%%%%%%%%%%%%%%%%%%%%%%%%%%%%%%%%%%%%%%%%%%%%%%%%%%%%%%%%%%%%%%%%%%%%%

%%%%%%%%%%%%%%%%%%%%%%%%%%%%%%%%%%%%%%%%%%%%%%%%%%%%%%%%%%%%%%%%%%%%%%%%%%%%%%%%

%%%%%%%%%%%%%%%%%%%%%%%%%%%%%%%%%%%%%%%%%%%%%%%%%%%%%%%%%%%%%%%%%%%%%%%%%%%%%%%%
%\section*{APPENDIX}


%\section*{ACKNOWLEDGMENT}

%We would like to thank Alessandro and Michael for the dataset collection.

%%%%%%%%%%%%%%%%%%%%%%%%%%%%%%%%%%%%%%%%%%%%%%%%%%%%%%%%%%%%%%%%%%%%%%%%%%%%%%%%

\bibliographystyle{IEEEtran}
\bibliography{root}

\end{document}